\documentclass[runningheads]{llncs}

\usepackage{eccv}

\usepackage{eccvabbrv}

\usepackage{graphicx}
\usepackage{pdfpages}
\usepackage{booktabs}

\usepackage{makecell}
\usepackage{amssymb}
\usepackage{multirow}
\usepackage{tabularx}

\definecolor{gain}{RGB}{34, 139, 34}
\definecolor{loss}{RGB}{205, 0, 0}

\newcommand{\resPos}[2]{\makecell{#1 ({\color{gain}+#2})}}
\newcommand{\resNeg}[2]{\makecell{#1 ({\color{loss}-#2})}}

\newcommand{\resPosBold}[2]{\makecell{\textbf{#1} ({\color{gain}\textbf{+#2}})}}

\usepackage[accsupp]{axessibility}  % Improves PDF readability for those with disabilities.

\usepackage[pagebackref,breaklinks,colorlinks,citecolor=eccvblue]{hyperref}
\usepackage{orcidlink}

\begin{document}

    % ---------------------------------------------------------------
    % DONE REVIEW: Replace with your title
    \title{3D-LENS: A 3D Lifting-based Elevated Novel-view Synthesis method for Single-View Aerial-Ground Re-Identification}

    % DONE REVIEW: If the paper title is too long for the running head, you can set
    % an abbreviated paper title here. If not, comment out.
    \titlerunning{3D-LENS}

    % TODO FINAL: Replace with your author list.
    % Include the authors' OCRID for the camera-ready version, if at all possible.
    \author{William Grolleau\inst{1,2}\orcidlink{0009-0003-8227-1866} \and
    Astrid Sabourin\inst{1} \and
    Guillaume Lapouge\inst{1}\orcidlink{0000-0002-4216-6696} \and
    Catherine Achard\inst{2}\orcidlink{0000-0002-5790-0830}}

    % TODO FINAL: Replace with an abbreviated list of authors.
    \authorrunning{W.~Grolleau et al.}
    % First names are abbreviated in the running head.
    % If there are more than two authors, 'et al.' is used.

    % TODO FINAL: Replace with your institution list.
    \institute{Universit\'e Paris-Saclay, CEA, List, F-91120 Palaiseau, France\\
    \email{\{william.grolleau,guillaume.lapouge\}@cea.fr} \and
    Sorbonne University, CNRS, Institute of Intelligent Systems and Robotics (ISIR), 4 Place Jussieu, 75005 Paris, France}

    \maketitle

    \begingroup\renewcommand\thefootnote{}\footnotetext{Accepted to the European Conference on Computer Vision (ECCV) 2026.}\endgroup

    \begin{abstract}
        Aerial-Ground Re-Identification (AG-ReID) is constrained by the viewpoint-domain gap, as drastic viewpoint disparities occlude or distort discriminative features, making cross-viewpoint image retrieval challenging.
        While existing methods rely on paired cross-view annotations, real-world deployments, such as wilderness search-and-rescue (SAR), often lack target-domain data, requiring retrieval from ground-level references alone.
        To our knowledge, we are the first to address this challenge by formalizing the \textbf{Single-View AG-ReID} (SV AG-ReID) setting, where models trained on a single real viewpoint must generalize to an unseen viewpoint.
        We propose \textbf{3D Lifting-based Elevated Novel-view Synthesis} (3D-LENS), a unified framework combining \textbf{geometrically-consistent novel view synthesis} that leverages large-scale 3D mesh reconstruction, with a \textbf{robust representation learning} scheme to mitigate synthetic-to-real bias.
        Unlike 2D generative baselines that suffer from geometric inconsistencies or prior 3D methods that are restricted to class-specific templates, our approach ensures view-consistent synthesis across diverse categories without predefined templates that fail to capture fine-grained details, such as carried objects.
        Extensive experiments demonstrate that our method achieves state-of-the-art performance on SV AG-ReID scenarios.
        Code and data will be released at \url{https://github.com/TurtleSmoke/3D-LENS}.
    \end{abstract}

    \section{Introduction}
    \label{sec:intro}

    Object Re-Identification (ReID) aims to retrieve identities across varying images in camera networks.

    Traditional ReID has primarily focused on view-homogeneous scenarios, such as CCTV ~\cite{zheng2015scalable, ristani2016performance} or Unmanned Aerial Vehicles (UAVs)~\cite{li2021uav, zhang2020person}.
    However, Aerial-Ground Re-Identification (AG-ReID)~\cite{nguyen2023aerial, nguyen2024ag} brings a novel challenge: discriminative features visible in oblique ground views are dissimilar or self-occluded in nadir aerial views.

\begin{figure}[tbp]
        \centering
        \includegraphics[width=0.8\linewidth]{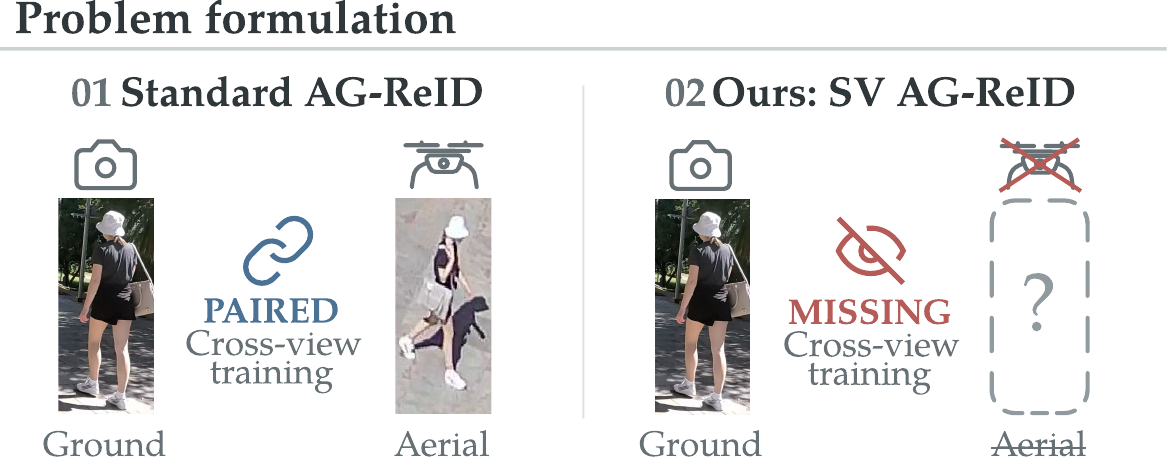}
        \caption{
            \textbf{Problem Formulation.}
            Standard AG-ReID (01) relies on paired data, which is often unavailable in real-world scenarios and leads to a missing cross-view challenge.
            To address this, we propose the Single-View Aerial Ground Re-Identification (SV AG-ReID) setting (02), where models must adapt from a single viewpoint to an unseen target viewpoint.
        }
        \label{fig:formulation}
    \end{figure}

    Existing AG-ReID methods predominantly rely on paired aerial-ground annotations~\cite{nguyen2023aerial, nguyen2024ag, Zhang_2024_CVPR, wang2025secap} or unsupervised adaptation~\cite{mei2025unsupervised} requiring target-viewpoint data.
    However, these assumptions are too restrictive for cold-start deployments such as search-and-rescue (SAR) missions, where an identity must be retrieved with UAVs using ground-level reference only.
    To overcome this dependency, we introduce the Single-View (SV AG-ReID) setting (\Cref{fig:formulation}).
    To our knowledge, this is the first approach that proposes to train models on a single real viewpoint (\eg, ground) and transfer to another (\eg, aerial).
    Unlike standard domain generalization ReID, which often targets stylistic shifts, such as lighting or weather~\cite{lee2025domain}, we focus on the viewpoint gap caused by elevation using explicit 3D geometry.

    Adapting standard ReID strategies to this setting is challenging.
    2D augmentations~\cite{chen2025towards, zhang2025dari, khalid2025bridging, li2026gsalign} fail to simulate the structural occlusion of large view shifts.
    Similarly, using generative models to synthesize missing viewpoints~\cite{niu2025synthesizing, zhang2025viperson, li2024vehiclegan, kim2024posediveposediversifiedaugmentationdiffusion} leads to geometric inconsistencies between novel views.
    In contrast, our approach leverages 3D reconstruction to lift the image into a fixed and textured mesh.
    Locking in both the geometry and texture allows us to render novel views that are inherently consistent with one another.

\begin{figure}[tbp]
        \centering
        \includegraphics[width=0.75\linewidth]{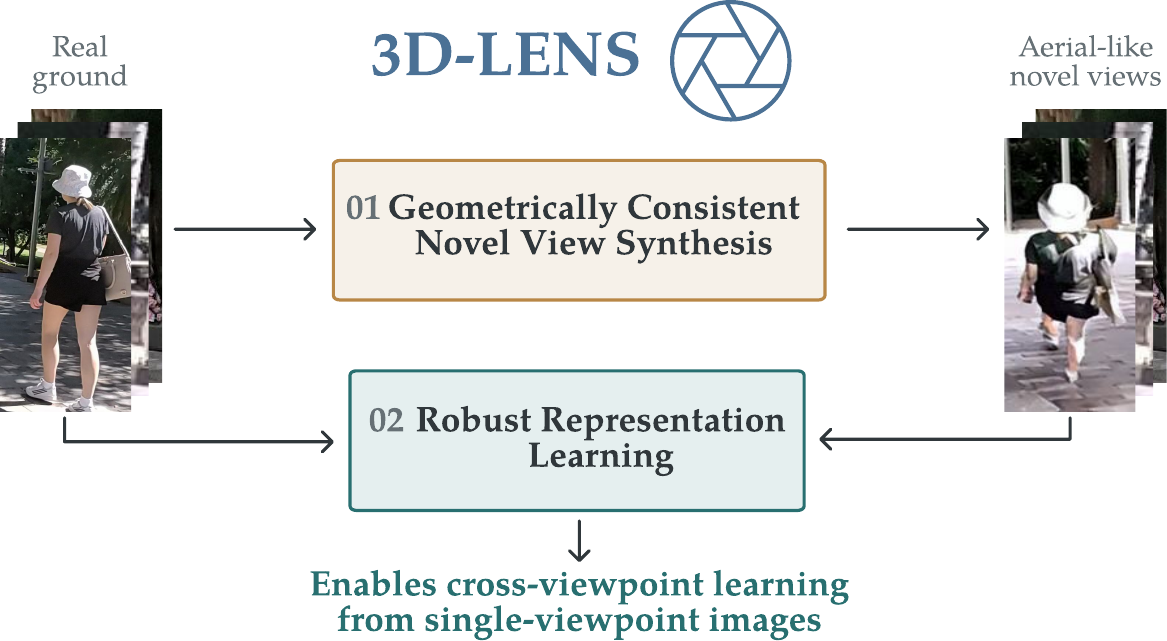}
        \caption{
            \textbf{Proposed Solution.}
            To tackle the SV AG-ReID, 3D-LENS uses a Geometrically Consistent Novel View Synthesis (01) to generate consistent novel views from single images using large-scale 3D synthesis, and a Robust Representation Learning scheme (02) to mitigate remaining biases in the synthetic data.
            This approach enables robust cross-viewpoint representation learning without requiring paired cross-view supervision.
        }
        \label{fig:overview}
    \end{figure}

    To tackle the SV AG-ReID scenario, we introduce 3D \textbf{L}ifting for \textbf{E}levated \textbf{N}ovel-view \textbf{S}ynthesis (\textbf{3D-LENS}) as illustrated in \Cref{fig:overview}.
    3D-LENS is a unified framework that combines geometrically-consistent novel view synthesis with a robust representation learning scheme.
    For the rendering phase, our method leverages recent advances in large-scale open-world 3D synthesis~\cite{zhao2025hunyuan3d} to reconstruct textured 3D meshes from single images and render view-consistent 2D images.
    Unlike previous geometric methods that rely on category-specific templates (\eg, SMPL~\cite{loper2023smpl} for humans or CAD~\cite{meng2022viewpoint} for vehicles), our approach is class-agnostic and exploits generic 3D priors learned from large-scale datasets.
    This enables the reconstruction of diverse categories, capturing fine-grained details, such as carried objects on pedestrians, while extending to any other class such as vehicles or animals.

    Crucially, lifting objects into 3D space decouples generation from rendering.
    This separation gives precise control over azimuth and elevation, allowing us to synthesize novel views consistent with the target viewpoint.
    To optimize joint training on both real and synthetic data, the second half of our framework employs our robust representation learning scheme.
    This uses curriculum scheduler to progressively bridge the viewpoint-domain gap, paired with a balanced sampling strategy to mitigate the synthetic-to-real bias.

    Our main contributions are as follows:
    \begin{itemize}
        \item To the best of our knowledge, \textbf{3D-LENS} is the first framework to address the \textbf{SV AG-ReID} setting.
        \item We pioneer the use of generic and large-scale 3D reconstruction in ReID with our geometrically-consistent novel view synthesis that generalize across diverse classes.
        \item We design a robust representation learning that uses a curriculum scheduler to overcome the viewpoint-domain gap, and a balanced sampling strategy to mitigate the synthetic-to-real bias.
        \item Experimental results demonstrate that our approach outperforms existing methods, surpassing the state-of-the-art by 14.30 percentage points (pp) mAP on AG-ReID and 21.20 pp mAP on AG-ReID.v2.
    \end{itemize}

    \section{Related works}
    \label{sec:related_works}

    \subsection{Aerial-Ground Re-Identification}

    Aerial-Ground Re-Identification (AG-ReID) addresses a challenging setup where ground and aerial domains exhibit drastic perspective differences.
    Early benchmarks, such as AG-ReID~\cite{nguyen2023aerial} and AG-ReID.v2~\cite{nguyen2024ag}, use annotated soft attributes (height, beard, glasses, \etc) to guide models toward view-invariant features.
    While effective, these methods require manual attribute annotation that limits scalability.

    To reduce annotation costs, recent approaches learn view-invariant features without explicit attribute supervision.
    For instance, VDT~\cite{Zhang_2024_CVPR} employs hierarchical attention and orthogonality constraints to disentangle view-related features from identity features.
    Building on this, SeCAP~\cite{wang2025secap} uses learnable prompts to refine identity representations.
    However, discarding view-specific information ignores identity features that are view-dependent, such as clothing designs or animal patterns.

    Rather than enforcing view-independent information, data-centric strategies use 2D augmentations to simulate viewpoint changes.
    Chen et al. \cite{chen2025towards} introduce learnable feature-level rotations, while DARI~\cite{zhang2025dari} employs rotated convolution kernels.
    Similarly, VIF~\cite{khalid2025bridging} proposes patch-level rotation, and GSAlign~\cite{li2026gsalign} models geometric deformations through keypoint-guided feature warping.
    Taking a generative approach, SD-ReID~\cite{wang2025sdreidviewawarestablediffusion} uses a diffusion model to synthesize identity features from opposing viewpoints.
    Without explicit geometric priors, 2D methods cannot simulate the structural gaps caused by self-occlusion in aerial views.

    These frameworks also rely on paired aerial-ground supervision to align features, which makes them ill-suited for SV AG-ReID, where the model must generalize from a single viewpoint without real cross-view images.

    We instead use a large 3D synthesis model~\cite{zhao2025hunyuan3d} to predict a textured 3D mesh from a single view.
    This explicit geometry guarantees accurate novel views rendering for any viewpoint, enabling reliable supervision for SV AG-ReID.

    \subsection{Generative View Synthesis in Re-Identification}

    Given the difficulty of acquiring large-scale datasets in ReID, synthetic data generation has become a popular strategy to augment training sets.

    Prior work constrains probabilistic generative models to improve structural realism.
    Methods, such as Diff-Person~\cite{niu2025synthesizing} and DCAC~\cite{li2025unleashing} condition generation on text or latent prompts.
    To improve controllability, VehicleGAN~\cite{li2024vehiclegan}, MVIGN \cite{xun2024multi} and VIPerson~\cite{zhang2025viperson} use 2D keypoints, while GCL~\cite{chen2021joint} and Pose-dIVE~\cite{kim2024posediveposediversifiedaugmentationdiffusion} utilize silhouettes or depth maps from 3D parametric models.
    However, these sparse 2D priors are insufficient to enforce strict 3D geometric constraints.
    Without an explicit 3D understanding, synthesizing multiple novel views often results in deformed individual shapes and textures.

    More critically, these approaches suffer from geometric cross-view inconsistencies.
    Since generative models synthesize pixels stochastically, they treat every novel view as independent.
    Even advanced conditional generation methods (\eg, ControlNet~\cite{zhang2023adding}, IP-Adapter~\cite{ye2023ip}) struggle to guarantee multi-view consistency.
    This independence introduces identity inconsistencies, such as a discriminative pattern shifting position or changing shape between two views of the same subject.

    Our approach achieves multi-view consistency by lifting a textured 3D mesh once, from which all subsequent views are rendered.
    Binding the texture to the geometry keeps identity features pixel-consistent across all synthesized angles, preventing identity inconsistencies and providing a reliable view signal for SV AG-ReID.

    \subsection{3D-Aware Re-Identification}

    Viewpoint variation is a central challenge in ReID~\cite{sun2019dissecting}, motivating the introduction of 3D geometric priors to disentangle identity from pose and camera perspective.

    Most 3D-aware methods rely on category-specific parametric templates, such as SMPL~\cite{loper2023smpl} for humans.
    3DSL~\cite{chen2021learning} regresses SMPL parameters to learn shape features, while 3DInvarReID~\cite{liu2023learning} uses a shape vector supervised by 3D ground truth.
    Moving beyond direct parameter regression, SEAS~\cite{zhu2024seas} uses distillation to guide an appearance encoder with implicit 3D features.
    With a multi-view approach, MV-3DSReID~\cite{yu2024pedestrian} aggregates multiple rendered views from SMPL.
    However, these methods often explicitly separate geometry and texture in their architecture, thereby weakening the relationship between the two.

    To overcome this separation, several approaches use point clouds to jointly model geometry and appearance.
    OG-Net~\cite{zheng2022parameter} uses Graph Neural Networks~\cite{scarselli2008graph} on SMPL point clouds, while PointReIDNet~\cite{wang20233d} further adds dynamic graph correction to mitigate shape distortions.
    Despite their robustness, point cloud extraction often suffers from sparsity and loss of high-frequency details.

    Alternative approaches focus on aligning textures to canonical views.
    For humans, methods like FusionTexReIDNet~\cite{nguyen2025beyond} unwrap body textures into UV maps via SMPL priors, while ClonedPerson~\cite{wang2022cloning} projects real-world textures onto 3D avatars.
    This geometric alignment extends to other domains, utilizing 3D CAD models for vehicles~\cite{meng2022viewpoint, tang2019pamtri} and UV unwrapping for animal pelage patterns~\cite{algasov2025unsupervised}.
    However, these parametric methods fail to capture external accessories (\eg, backpacks) and cannot generalize to classes lacking predefined 3D templates.

    Consequently, 3D-aware methods share a common limitation: they rely on category-specific priors.
    In contrast, our approach leverages class-agnostic 3D reconstruction to render consistent novel views for any object category.
    By synthesizing views with explicit 3D transformations on a mesh, we unify geometry and appearance to learn view-agnostic features regardless of the object category.

    \section{Method}

\begin{figure}[tbp]
        \centering
        \includegraphics[width=1.0\linewidth]{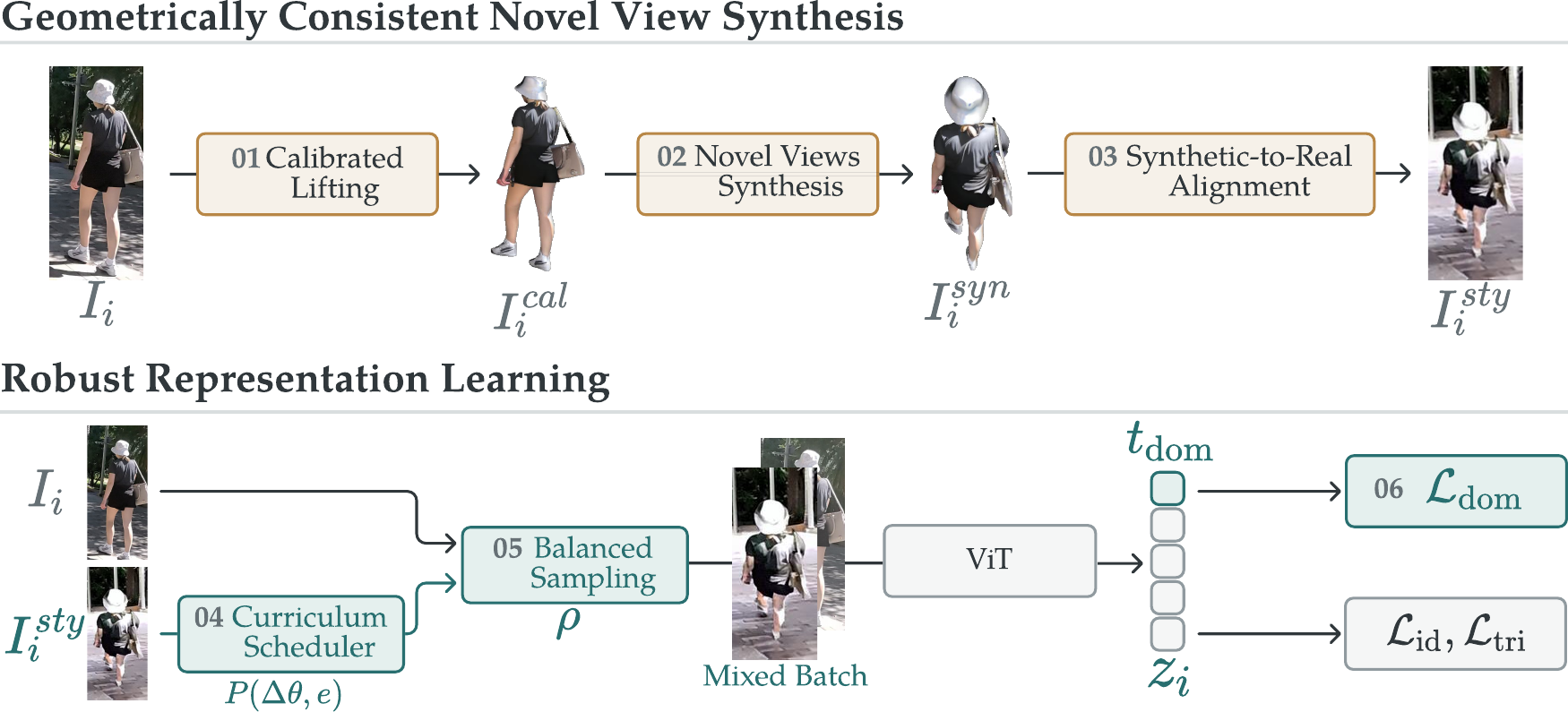}
        \caption{
            Overview of the 3D-LENS framework.
            For the Geometrically Consistent Novel View Synthesis, a source image $I_i$ is canonically lifted to a 3D representation (01), followed by novel view synthesis (02) and synthetic-to-real alignment (03) to produce target-view images $I_i^{sty}$.
            In our Robust Representation Learning scheme, an elevation-based curriculum scheduler (04) progressively introduces these synthetic views into training, then a balanced sampling strategy (05) combines the synthetic views $I_i^{sty}$ and real images $I_i$ to form a mixed batch, which is optimized using standard reid losses and a domain loss (06).
        }
        \label{fig:lens}
    \end{figure}

    3D-LENS (\Cref{fig:lens}) addresses SV AG-ReID through two core contributions: (i) \textbf{Geometrically-Consistent Novel View Synthesis}, which generates consistent novel views from single images using large-scale 3D synthesis, and (ii) a \textbf{Robust Representation Learning} scheme, which employs curriculum scheduler to progressively learn the viewpoint shift from the generated novel views and a balanced sampling to mitigate the synthetic-to-real bias.

    \subsection{Problem Formulation}

    Standard AG-ReID typically assumes a fully paired training set $\mathcal{D} = \{ \mathcal{D}_A, \mathcal{D}_G \}$, representing aerial and ground viewpoints respectively.
    However, in practical deployment scenarios such as search-and-rescue or wildlife monitoring, acquiring paired cross-view data is often logistically impractical.
    Under these conditions, a model must generalize from a single available viewpoint to a completely unseen target viewpoint.

    To address this, we define the Single-View AG-ReID (SV AG-ReID) setting.
    Given a labeled source viewpoint $\mathcal{D}_{src}$ containing images $I$ from a single viewpoint (\eg, ground $\mathcal{D}_G$), the objective is to learn a robust representation $\mathcal{F}(I)$ that generalizes to an unseen target viewpoint $\mathcal{D}_{t}$ (\eg, aerial $\mathcal{D}_A$).

    To the best of our knowledge, we are the first to formalize and tackle the SV AG-ReID task.
    The core difficulty is the lack of cross-view supervision: the model cannot observe the geometric and appearance changes induced by viewpoints shift during training.
    We solve this issue by explicitly modeling the 3D geometry of the source data, simulating the target viewpoint-domain without requiring real-world target samples.

    \subsection{Geometrically-Consistent Novel View Synthesis}
    \label{subsec:geometrically-consistent-novel-view-synthesis}

    Simulating cross-view data help models to learn robust representations in SV AG-ReID.
    However, 2D generative models often hallucinate inconsistent textures across independent views, while existing 3D-aware methods rely on category-specific templates that discard instance-specific details, such as accessories and fail to generalize to novel classes.

    To overcome these limitations, we propose a geometrically-consistent pipeline for novel view synthesis illustrated in \Cref{fig:lens}.
    First we leverage a large-scale 3D synthesis model to lift a single image into a textured mesh.
    Next, We optimize the virtual camera pose to align the rendered mesh with the original perspective, allowing us to synthesize geometrically-consistent novel views by applying controlled perturbations to this calibrated pose.
    Finally, we bridge the synthetic-to-real domain gap by compositing the rendered foregrounds onto real backgrounds and applying style transfer to align their visual characteristics, preventing the network from overfitting to artificial rendering artifacts.

    \subsubsection*{2D-to-3D Calibrated Lifting and Alignment.}

    Given a source image $I_i \in \mathcal{D}_{src}$, we isolate the foreground identity by extracting a mask $M_i$ using an off-the-shelf segmentation model (\eg, YOLOv11~\cite{khanam2024yolov11}).
    A single-view reconstruction estimator $\mathcal{F}_{3D}$ (\eg, Hunyuan3D~\cite{zhao2025hunyuan3d}) then lifts the 2D signal into a textured mesh $\mathcal{M}_{i}$, unifying geometry and appearance.

    Because $\mathcal{F}_{3D}$ estimates meshes in an arbitrary coordinate space, the default virtual camera may not match the original perspective of $I_i$.
    We therefore optimize the virtual camera pose $\Pi_{i} = [R_i | t_i]$ to align the rendered mesh with the reference image.

    We define $\mathcal{R}(\mathcal{M}_i, \Pi)$ as a rendering function producing foreground-only images of $\mathcal{M}_i$ under camera parameters~$\Pi$.
    The calibrated pose $\Pi_i$ is recovered by maximising the Intersection over Union (IoU) between observed Mask $M_i$ and the binary silhouette of the rendered image $\mathcal{S}(\mathcal{R}(\mathcal{M}_i, \Pi))$:

    \begin{equation}
        \Pi_{i} = \arg \max_{\Pi} \text{IoU}\left(M_{i}, \mathcal{S}(\mathcal{R}(\mathcal{M}_i, \Pi))\right).
    \end{equation}

    To keep only high-fidelity data, we discard meshes if the optimal IoU falls below a threshold $\tau_{iou}$.
    The calibrated pose then yields a reference rendering $I_i^{cal} = \mathcal{R}(\mathcal{M}_i, \Pi_i)$, which serves as the starting point for novel-view synthesis.

    \subsubsection*{Geometrically-Consistent View Synthesis.}

    We synthesize a novel view $I_i^{syn}$ by perturbing the calibrated pose $\Pi_i$ using the aligned mesh $\left < \mathcal{M}_i, \Pi_i \right >$.
    We shift the elevation by $\Delta\theta$ and azimuth by $\Delta\phi$ to define a new camera pose $\tilde{\Pi}_i$, yielding the rendered image $I_i^{syn} = \mathcal{R}(\mathcal{M}_i, \tilde{\Pi}_i)$.

    Because the mesh is inferred from a single view, it may contain hallucinated textures in self-occluded or unseen regions, so we restrict our synthesis angles to preserve visual fidelity.
    Specifically, we constrain elevation shifts toward the nadir for ground-to-aerial synthesis, and toward the horizon for aerial-to-ground synthesis.

    This controlled rendering bridges the SV AG-ReID gap by providing geometrically consistent cross-view supervision while preserving the identity of the source image $I_i$.
    Depending on the task, it explicitly simulates the missing target domain by synthesizing aerial perspectives from ground images (G$\rightarrow$A), or vice versa (A$\rightarrow$G).

    \subsubsection*{Synthetic-to-Real Domain Alignment.}

    Because the renderer $\mathcal{R}$ outputs foreground-only projections, the resulting views lack natural scene context.
    We close this appearance gap in two steps.
    First, we composite the rendered foreground onto a real background, then we apply style transfer to harmonize the synthetic foreground with the real background.

    First, we generate a clean background $\mathcal{B}_j$ by masking the foreground of a randomly chosen training image $I_j$ ($j \neq i$) and applying an inpainting model (\eg, LaMa~\cite{suvorov2022resolutionrobustlargemaskinpainting}).
    Then, we composite the rendered foreground $I_i^{syn}$ onto the real background to obtain $I_i^{comp} = I_i^{syn} \odot M_i + \mathcal{B}_j \odot (1 - M_i)$, where $\odot$ denotes element-wise multiplication.

    However, naive compositing introduces boundary artifacts and illumination inconsistencies between the synthetic subject and the photographic background.
    To resolve this discrepancy, we apply a style transfer module (\eg, StyleID~\cite{chung2024styleid}) to align the visual appearance of the composite image $I_i^{comp}$ with a randomly sampled real image from $\mathcal{D}_{src}$ and get our final synthetic view $I_i^{sty}$.
    These steps ensure that the synthetic views visually match the real training distribution and reduce the risk of the model overfitting to rendering artifacts.

    \subsection{Robust Representation Learning}

    \subsubsection*{Curriculum Scheduler}

    We use curriculum learning to schedule sample difficulty over training.
    Our pipeline makes this straightforward because elevation shifts $|\Delta\theta|$ provide a direct measure of sample difficulty, without relying on external heuristics.

    During training, a synthetic sample generated at epoch $e$ is rejected with probability $P(\Delta\theta, e)$:

    \begin{equation}
        P(\Delta\theta, e) = \max \left( 0,\left( 1 - \frac{e}{E_{\text{warmup}}} \right) \right) \frac{|\Delta\theta|}{|\Delta\theta_{\max}|}
    \end{equation}

    where we define the maximum synthesis angle as $|\Delta\theta_{\max}| = 30^\circ$, and the warmup period as $E_{\text{warmup}} = 20$ epochs in our experiments.

    Rejection probability $P$ grows linearly with the magnitude of the elevation shift $|\Delta\theta|$: near-source views are always accepted, while extreme viewpoints have high initial rejection rates.
    As training progresses, the global rejection penalty decreases linearly to zero at epoch $E_{\text{warmup}}$, after which all synthetic samples are used without restriction.

    This allows the model to reliably learn identity features from stable viewpoints before encountering the full elevation gap.

    \subsubsection*{Balanced Sampling}

    Standard ReID training~\cite{luo2019bag} samples mini-batches of $P$ identities with $K$ instances each.
    In our setting, each real image $I_i \in \mathcal{D}_{src}$ can produce several synthetic views, so $|\mathcal{D}_{syn}| \gg |\mathcal{D}_{src}|$.
    Naive uniform sampling over $\mathcal{D}_{src} \cup \mathcal{D}_{syn}$ results in imbalanced batches dominated by synthetic data during training and biases the model toward synthetic artifacts.

    We prevent this bias by enforcing a fixed ratio $\rho$ of synthetic and real samples such that $K = \rho K_{real} + (1-\rho) K_{syn}$ regardless of the number of synthetic views.
    One training epoch is defined as a complete pass over $\mathcal{D}_{src}$, so every real image is seen exactly once per epoch.
    At each iteration, synthetic samples are drawn to fill the $K_{syn}$ slots, and any unused views are sampled in subsequent epochs, so the full synthetic set is used over training.

    \subsubsection*{Losses}

    To further reduce the synthetic-to-real gaps, we introduce a domain classification loss that encourages the model to separate domain-specific cues from identity features.
    We introduce a domain token $t_{\text{dom}}$ to the ViT~\cite{dosovitskiy2021image} backbone and supervise it with a domain classification loss $\mathcal{L}_{\text{dom}}$ that predicts whether the input is from $\mathcal{D}_{src}$ or $\mathcal{D}_{syn}$.

    For identity supervision, we follow standard baseline~\cite{luo2019bag} and combine cross-entropy $\mathcal{L}_{\text{id}}$ and triplet $\mathcal{L}_{\text{tri}}$ losses.
    The full objective is:

    \begin{equation}
        \mathcal{L}_{\text{total}} = \lambda_{\text{id}} \mathcal{L}_{\text{id}} + \lambda_{\text{tri}} \mathcal{L}_{\text{tri}} + \lambda_{\text{dom}} \mathcal{L}_{\text{dom}}
    \end{equation}

    where $\lambda_{\text{id}}$, $\lambda_{\text{tri}}$, and $\lambda_{\text{dom}}$ are scalar weights for each term.

    \section{Experiments}

    \subsection*{Experimental Settings}

    We evaluate our method on two Person AG-ReID benchmarks: AG-ReID~\cite{nguyen2023aerial} and AG-ReID.v2~\cite{nguyen2024ag}.
    AG-ReID contains 21,983 images of 388 identities captured by one ground camera and one UAV.
    AG-ReID.v2 has 100,502 images of 1,615 identities across three platforms: UAVs, wearable cameras, and CCTV.

    We also evaluate on MOO~\cite{grolleau2026moomultivieworientedobservations}, a synthetic AG-ReID dataset for cattle, to test whether our approach transfers beyond pedestrians.
    MOO contains 128,000 images of 1,000 identities rendered at elevations ranging from $-25^\circ$ to $90^\circ$.

    Following prior work~\cite{luo2019bag}, we use mean Average Precision (mAP) and Rank-1 accuracy as evaluation metrics, computed using the standard Cumulative Matching Characteristic (CMC) curve.

    \subsubsection*{Evaluation Protocols}

    \begin{table}[tbp]
        \caption{
            \textbf{State-of-the-Art Comparison on AG-ReID Dataset.}
            Training is performed on a single viewpoint (Aerial or Ground) and evaluated in a cross-view Query $\rightarrow$ Gallery scenario on Aerial (A), and Ground (G).
            Results are reported as mAP and Rank-1 (\%).
            Best results are in \textbf{bold}, and second-best are \underline{underlined}.
        }
        \label{tab:sota_agreid}
        \centering

        \scriptsize
        \setlength{\tabcolsep}{6pt}

        \begin{tabular}{l c cccc cccc}
            \toprule
            \multicolumn{10}{c}{\textbf{AG-ReID}} \\

            \cmidrule(lr){1-10}

            &
            & \multicolumn{4}{c}{\textbf{Aerial Training}}
            & \multicolumn{4}{c}{\textbf{Ground Training}} \\

            \cmidrule(lr){3-6} \cmidrule(lr){7-10}

            \textbf{Methods} & \textbf{Year}
            & \multicolumn{2}{c}{$A \rightarrow G$}
            & \multicolumn{2}{c}{$G \rightarrow A$}
            & \multicolumn{2}{c}{$A \rightarrow G$}
            & \multicolumn{2}{c}{$G \rightarrow A$} \\

            \cmidrule(lr){3-4} \cmidrule(lr){5-6} \cmidrule(lr){7-8} \cmidrule(lr){9-10}

            & & mAP & R-1 & mAP & R-1
            & mAP & R-1 & mAP & R-1 \\
            \midrule

            BoT~\cite{luo2019bag} & 2019
            & 41.7 & 55.7 & 29.6 & 34.3
            & 45.3 & 51.8 & 49.6 & 60.2 \\

            ViT~\cite{dosovitskiy2021image} & 2020
            & 42.6 & 58.1 & 31.9 & 35.8
            & 46.9 & 54.6 & 49.8 & 60.1 \\

            AGW~\cite{ye2021deep} & 2021
            & 41.4 & 55.3 & 32.2 & 36.7
            & 42.3 & 51.0 & 47.3 & 59.5 \\

            TransReID~\cite{he2021transreid} & 2021
            & 32.7 & 46.4 & 23.5 & 25.2
            & 44.2 & 50.7 & 49.7 & 62.6 \\

            RotTrans~\cite{chen2022rotation} & 2022
            & 35.2 & 50.2 & 23.2 & 35.8
            & 44.8 & 50.3 & 48.8 & 59.3 \\

            PASS~\cite{zhu2022pass} & 2022
            & 39.1 & 52.2 & 26.0 & 28.6
            & 47.7 & 56.2 & 47.8 & 58.0 \\

            SBS~\cite{he2023fastreid} & 2023
            & 16.8 & 27.9 & 15.5 & 15.2
            & 27.1 & 32.3 & 27.5 & 37.5 \\

            DCT~\cite{li2023dc} & 2023
            & 22.0 & 35.8 & 14.0 & 14.2
            & 36.2 & 43.8 & 40.6 & 52.3 \\

            VDT~\cite{Zhang_2024_CVPR} & 2024
            & 39.0 & 53.3 & 31.7 & 36.1
            & 48.3 & 56.0 & 51.4 & 63.2 \\

            DCAC~\cite{li2025unleashing} & 2025
            & \underline{43.4} & \underline{60.0} & \underline{35.2} & \underline{42.7}
            & \underline{49.7} & \underline{61.6} & \underline{52.9} & \underline{66.4} \\

            \midrule

            \textbf{3D-LENS} & \textbf{2026}
            & \textbf{49.8} & \textbf{62.9} & \textbf{49.5} & \textbf{57.7}
            & \textbf{53.3} & \textbf{65.2} & \textbf{55.1} & \textbf{67.4} \\

            \bottomrule
        \end{tabular}
    \end{table}

    We define the SV AG-ReID evaluation protocol as follows.
    Models are trained exclusively on a single viewpoint-domain and evaluated on cross-domain query-to-gallery matching.

    For AG-ReID and MOO, there are two training configurations.
    The first is trained on Aerial (A) and Ground (G) images, while the second uses Top (T) and Side (S) images.
    Both models are then evaluated in a cross-view Query $\rightarrow$ Gallery scenario (\ie $A\rightarrow G$ and $G\rightarrow A$), yielding four scenarios per dataset.

    AG-ReID.v2 introduces two ground-level platforms (CCTV (C) and Wearable (W)), so we evaluate three training configurations: Aerial (A), CCTV (C), and Wearable (W).
    Models trained on $A$ are tested against both $C$ and $W$ ($A\rightarrow C$, $C\rightarrow A$, $A\rightarrow W$, $W\rightarrow A$), and models trained on $C$ or $W$ are each tested against $A$ ($C\rightarrow A$, $A\rightarrow C$ and $W\rightarrow A$, $A\rightarrow W$), for a total of eight evaluation scenarios.

    \subsubsection*{Implementation Details}

    We build upon BoT~\cite{luo2019bag} and use a ViT~\cite{dosovitskiy2021image} base as our backbone and baseline.
    In all experiments, ViT is the baseline upon which each of our contributions is added.
    For a fair comparison we use a ViT base and standardize the input size and learning rate across all competing methods, except when a method's original baseline explicitly requires specific hyperparameter settings.
    We use Yolov11~\cite{khanam2024yolov11} for segmentation and Hunyuan3D~\cite{zhao2025hunyuan3d} for 3D reconstruction.
    We use $\tau = 0.7$ for IoU-based mesh filtering and StyleID~\cite{chung2024styleid} for style transfer.
    Elevation shifts are constrained to $\Delta\theta \in [0^\circ, 30^\circ]$ for ground-to-aerial synthesis and $\Delta\theta \in [-30^\circ, 0^\circ]$ for aerial-to-ground, and azimuth shifts are limited to $|\Delta\phi| \le 30^\circ$, an ablation of this range is provided in the supplementary material.
    $\Delta\theta$ is uniformly sampled for the elevations, while $\Delta\phi$ is randomly sampled within the defined range.
    For the curriculum, we use $E_{\text{warmup}} = 20$ epochs and $\Delta\theta_{\max} = 30^\circ$.
    Mini-batches contain $P=32$ identities with $K=4$ images each ($K_{real}=2$, $K_{syn}=2$), giving a 1:1 real-to-synthetic ratio and a total batch size of 128.
    Images are resized to $256\times128$ and augmented with random cropping, horizontal flipping, and random erasing.
    We train for 120 epochs with SGD (weight decay$=10^{-4}$, momentum$=0.9$) with a lr$=8\times10^{-4}$ for AG-ReID and AG-ReID.v2, and lr$=8\times10^{-3}$ for MOO with cosine scheduling.
    Loss weights are $\lambda_{\text{id}} = 1.0$, $\lambda_{\text{tri}} = 1.0$, and $\lambda_{\text{dom}} = 0.5$.
    The full data generation pipeline incurs approximately 272 A100 and 8 P5000 GPU-hours on AG-ReID, complete cost breakdown is provided in the supplementary material.

    \subsection{Comparison with State-of-the-Art Methods}

    We compare our proposed 3D-LENS against two real-world AG-ReID datasets: AG-ReID (\Cref{tab:sota_agreid}), and AG-ReID.v2 (\Cref{tab:sota_agreid2_aerial,tab:sota_agreid2_cctv_wearable}).
    The compared methods include strong ReID baselines (\eg, BoT~\cite{luo2019bag}, TransReID~\cite{he2021transreid}), viewpoint-agnostic approaches (VDT~\cite{Zhang_2024_CVPR}, RotTrans~\cite{chen2022rotation}), large-scale pretraining models (PASS~\cite{zhu2022pass}), and recent generative adaptation methods (DCAC~\cite{li2025unleashing}).
    For methods built on TransReID, we omit the Side Information Embedding (SIE) module from the standard TransReID architecture as SIE fundamentally relies on explicit multi-view inputs and cannot work in the single-view setting.
    All methods use a ViT base backbone and are pretrained on ImageNet-21k~\cite{ridnik2021imagenet} except for PASS, which uses its own large-scale pretraining.

    \begin{table}[tbp]
        \caption{
            \textbf{State-of-the-Art Comparison on AG-ReID.v2 Dataset.}
            Training is performed on a single viewpoint (Aerial) and evaluated in a cross-view Query $\rightarrow$ Gallery scenario against CCTV (C) and Wearable (W).
            Results are reported as mAP and Rank-1 (\%).
            Best results are in \textbf{bold}, and second-best are \underline{underlined}.
        }
        \label{tab:sota_agreid2_aerial}
        \centering

        \scriptsize
        \setlength{\tabcolsep}{6pt}

        \begin{tabular}{l c cccc cccc}
            \toprule
            \multicolumn{10}{c}{\textbf{AG-ReID.v2}} \\

            \cmidrule(lr){1-10}

            &
            & \multicolumn{4}{c}{\textbf{Aerial Training}}
            & \multicolumn{4}{c}{\textbf{Aerial Training}} \\

            \cmidrule(lr){3-6} \cmidrule(lr){7-10}

            \textbf{Methods} & \textbf{Year}
            & \multicolumn{2}{c}{$C \rightarrow A$}
            & \multicolumn{2}{c}{$A \rightarrow C$}
            & \multicolumn{2}{c}{$W \rightarrow A$}
            & \multicolumn{2}{c}{$A \rightarrow W$} \\

            \cmidrule(lr){3-4} \cmidrule(lr){5-6} \cmidrule(lr){7-8} \cmidrule(lr){9-10}

            & & mAP & R-1 & mAP & R-1
            & mAP & R-1 & mAP & R-1 \\
            \midrule

            BoT~\cite{luo2019bag} & 2019
            & 14.4 & 16.8 & 32.5 & 42.4
            & 18.8 & 21.2 & 28.7 & 40.5 \\

            ViT~\cite{dosovitskiy2021image} & 2020
            & 14.7 & 16.7 & 33.2 & 43.1
            & 19.1 & 21.5 & 27.4 & 38.1 \\

            AGW~\cite{ye2021deep} & 2021
            & 14.6 & 17.8 & 33.2 & 42.6
            & 19.9 & \underline{23.9} & 30.2 & 40.9 \\

            TransReID~\cite{he2021transreid} & 2021
            & 11.5 & 12.4 & 25.3 & 34.0
            & 15.8 & 18.4 & 23.8 & 34.1 \\

            RotTrans~\cite{chen2022rotation} & 2022
            & 15.2 & 16.6 & 32.6 & 42.1
            & \underline{20.7} & 21.8 & 30.4 & 42.5 \\

            PASS~\cite{zhu2022pass} & 2022
            & \underline{16.6} & \underline{21.1} & 32.4 & 43.6
            & 16.2 & 19.4 & 26.3 & 38.8 \\

            SBS~\cite{he2023fastreid} & 2023
            & 2.5 & 2.7 & 3.8 & 5.1
            & 3.2 & 3.3 & 2.7 & 2.8 \\

            DCT~\cite{li2023dc} & 2023
            & 5.8 & 6.2 & 12.6 & 20.2
            & 10.4 & 12.1 & 12.8 & 22.1 \\

            VDT~\cite{Zhang_2024_CVPR} & 2024
            & 14.6 & 17.3 & 32.5 & 42.5
            & 18.3 & 21.4 & 26.4 & 37.6 \\

            DCAC~\cite{li2025unleashing} & 2025
            & 14.4 & 18.1 & \underline{34.9} & \underline{47.9}
            & 19.8 & 23.5 & \underline{32.6} & \underline{46.8} \\

            \midrule

            \textbf{3D-LENS}
            & \textbf{2026} & \textbf{37.8} & \textbf{47.1} & \textbf{42.8} & \textbf{55.7}
            & \textbf{28.6} & \textbf{35.8} & \textbf{34.3} & \textbf{47.1} \\

            \bottomrule
        \end{tabular}
    \end{table}

    As shown in \Cref{tab:sota_agreid}, 3D-LENS outperforms SoTA performance across all cross-view scenarios.
    Notably, the results reveal clear limitations of existing methods in the SV AG-ReID setting.
    Current viewpoint-agnostic methods struggle to handle strong cross-view scenarios under single-view training constraints (\eg, VDT, RotTrans), while large-scale pretraining models (\eg, PASS) fail to generalize to such extreme elevation changes.
    Generative adaptation methods (\eg, DCAC) prove to be the most competitive among prior arts, yet they still fall short of our approach.
    For example, in the Aerial training and Ground-to-Aerial ($G \rightarrow A$) setting, 3D-LENS achieves a mAP of $49.5\%$, outperforming DCAC, by $+14.3$ percentages points (pp) and the ViT baseline by $+17.6$ pp.
    This confirms that our precise, geometrically-consistent novel view synthesis is significantly more effective than purely data-driven appearance mapping.

    \Cref{tab:sota_agreid2_aerial} and \Cref{tab:sota_agreid2_cctv_wearable} evaluate methods on the more complex AG-ReID.v2 dataset, encompassing Aerial, CCTV, and Wearable camera views.
    In these configurations, large-scale pre-training models like PASS emerge as the strongest existing baselines, closely followed by DCAC and RotTrans.
    However, 3D-LENS still exceeds all prior methods by a significant margin, in the CCTV-to-Aerial ($C \rightarrow A$) scenario under Aerial training (\Cref{tab:sota_agreid2_aerial}), 3D-LENS achieves a mAP of $37.8\%$ and outperforms PASS by $21.2$ pp and the ViT baseline by $23.1$ pp.
    Consistent improvements on all other training and cross-view scenarios confirm our method's ability to learn view-invariant representations.
    To further validate the generalizability of our approach, we evaluate our data generation on the MOO dataset (\Cref{tab:sota_moo}), which features cross-view re-identification between top-down and side views.
    In this configuration, rotation invariance plays a crucial role and the best existing method is thus RotTrans~\cite{chen2022rotation} with its explicit rotation-equivariant architecture.
    Specifically, in the Top training and Top-to-Side ($T \rightarrow S$) scenario, 3D-LENS achieves a mAP of $51.4\%$, outperforming RotTrans by $+10.0$ pp and the ViT baseline by $+14.9$ pp.
    This demonstrates the effectiveness of our novel view synthesis and robust representation learning in bridging extreme viewpoint gaps across different classes.

    \begin{table}[tbp]
        \caption{
            \textbf{State-of-the-Art Comparison on AG-ReID.v2 Dataset.}
            Training is performed on a single viewpoint (CCTV or Wearable) and evaluated in a cross-view Query $\rightarrow$ Gallery scenario against Aerial (A).
            Results are reported as mAP and Rank-1 (\%).
            Best results are in \textbf{bold}, and second-best are \underline{underlined}.
        }
        \label{tab:sota_agreid2_cctv_wearable}
        \centering

        \scriptsize
        \setlength{\tabcolsep}{6pt}

        \begin{tabular}{l c cccc cccc}
            \toprule
            \multicolumn{10}{c}{\textbf{AG-ReID.v2}} \\

            \cmidrule(lr){1-10}

            &
            & \multicolumn{4}{c}{\textbf{CCTV Training}}
            & \multicolumn{4}{c}{\textbf{Wearable Training}} \\

            \cmidrule(lr){3-6} \cmidrule(lr){7-10}

            \textbf{Methods} & \textbf{Year}
            & \multicolumn{2}{c}{$C \rightarrow A$}
            & \multicolumn{2}{c}{$A \rightarrow C$}
            & \multicolumn{2}{c}{$W \rightarrow A$}
            & \multicolumn{2}{c}{$A \rightarrow W$} \\

            \cmidrule(lr){3-4} \cmidrule(lr){5-6} \cmidrule(lr){7-8} \cmidrule(lr){9-10}

            & & mAP & R-1 & mAP & R-1
            & mAP & R-1 & mAP & R-1 \\
            \midrule

            BoT~\cite{luo2019bag} & 2019
            & 42.3 & 52.6 & 34.5 & 40.0
            & 31.6 & 38.0 & 25.1 & 28.9 \\

            ViT~\cite{dosovitskiy2021image} & 2020
            & 41.8 & 51.8 & 35.0 & 41.0
            & 31.0 & 37.1 & 23.9 & 27.9 \\

            AGW~\cite{ye2021deep} & 2021
            & 38.3 & 47.4 & 31.5 & 38.2
            & \underline{32.4} & 39.1 & 25.3 & 31.2 \\

            TransReID~\cite{he2021transreid} & 2021
            & 36.6 & 47.4 & 31.1 & 37.3
            & 26.0 & 33.7 & 20.7 & 23.8 \\

            RotTrans~\cite{chen2022rotation} & 2022
            & 41.6 & 52.4 & 34.8 & 39.7
            & 31.6 & 40.4 & 28.7 & 31.2 \\

            PASS~\cite{zhu2022pass} & 2022
            & 40.5 & 50.5 & 36.5 & 42.4
            & 31.0 & 39.8 & 25.8 & 29.8 \\

            SBS~\cite{he2023fastreid} & 2023
            & 14.3 & 21.5 & 12.6 & 14.3
            & 3.8 & 5.0 & 3.5 & 3.0 \\

            DCT~\cite{li2023dc} & 2023
            & 31.6 & 43.4 & 26.0 & 30.6
            & 17.9 & 26.6 & 15.8 & 17.8 \\

            VDT~\cite{Zhang_2024_CVPR} & 2024
            & 41.1 & 51.7 & 33.9 & 39.8
            & 29.6 & 37.2 & 23.3 & 26.4 \\

            DCAC~\cite{li2025unleashing} & 2025
            & \underline{45.6} & \underline{57.9} & \underline{39.0} & \underline{48.4}
            & 31.9 & \underline{41.6} & \underline{30.1} & \underline{35.8} \\

            \midrule

            \textbf{3D-LENS} & \textbf{2026}
            & \textbf{47.2} & \textbf{58.5} & \textbf{41.4} & \textbf{50.8}
            & \textbf{38.3} & \textbf{46.2} & \textbf{38.9} & \textbf{47.9} \\

            \bottomrule
        \end{tabular}
    \end{table}

    The results validate that our method effectively bridge the complex viewpoint gap in SV AG-ReID with its explicit 3D geometric consistency and precise novel viewpoint synthesis, an aspect that standard feature alignment, pretraining, or 2D generation do not yet achieve.

    \subsection{Ablation Studies}
    \label{subsec:ablation-studies}

    To validate each component introduced in our method, we conduct ablation studies on the AG-ReID dataset.
    \Cref{tab:ablation_ground_full} reports the individual contributions of each module on the Ground training set.
    Due to space constraints, we present a representative subset of combinations here.
    Complete results and hyperparameter ablations are provided in the supplementary material.

    Interestingly, naively introducing novel view synthesis (Index 1) in isolation degrades performance across both settings (\eg, $-3.2$ mAP pp on $A \rightarrow G$ and $-1.5$ mAP pp on $G \rightarrow A$).
    This initial drop confirms that the raw images harm representation learning.
    However, balanced sampling (Index 5) or style transfer (Index 3) largely resolves this issue.
    Integrating background compositing over style transfer (Index 3 vs Index 4) further improves performance by closing the appearance gap.
    Specifically, these geometrically-consistent views (Index 4) yield improvements over the baselines (\eg, $+4.0$ mAP pp on $A \rightarrow G$ and $+1.8$ mAP pp on $G \rightarrow A$).

    \begin{table}[tbp]
        \caption{
            \textbf{State-of-the-Art Comparison on MOO.}
            Training is performed on a single viewpoint (Top or Side) and evaluated in a cross-view Query $\rightarrow$ Gallery scenario on Top (T) and Side (S) views.
            Results are reported as mAP and Rank-1 (\%).
            Best results are in \textbf{bold}, and second-best are \underline{underlined}.
        }
        \label{tab:sota_moo}
        \centering

        \scriptsize
        \setlength{\tabcolsep}{6pt}

        \begin{tabular}{l c cccc cccc}
            \toprule
            \multicolumn{10}{c}{\textbf{MOO}} \\

            \cmidrule(lr){1-10}

            &
            & \multicolumn{4}{c}{\textbf{Top Training}}
            & \multicolumn{4}{c}{\textbf{Side Training}} \\

            \cmidrule(lr){3-6} \cmidrule(lr){7-10}

            \textbf{Methods} & \textbf{Year}
            & \multicolumn{2}{c}{$T \rightarrow S$}
            & \multicolumn{2}{c}{$S \rightarrow T$}
            & \multicolumn{2}{c}{$T \rightarrow S$}
            & \multicolumn{2}{c}{$S \rightarrow T$} \\

            \cmidrule(lr){3-4} \cmidrule(lr){5-6} \cmidrule(lr){7-8} \cmidrule(lr){9-10}

            & & mAP & R-1 & mAP & R-1
            & mAP & R-1 & mAP & R-1 \\
            \midrule

            BoT~\cite{luo2019bag} & 2019
            & 37.1 & 95.9 & 39.1 & 43.5
            & 19.3 & 43.1 & 19.4 & 94.2 \\

            ViT~\cite{dosovitskiy2021image} & 2020
            & 36.5 & 93.9 & 37.8 & 41.6
            & 19.8 & 43.5 & 20.0 & 94.8 \\

            AGW~\cite{ye2021deep} & 2021
            & 36.1 & 89.2 & 37.8 & 44.0
            & 13.5 & 40.8 & 13.8 & 76.6 \\

            TransReID~\cite{he2021transreid} & 2021
            & 37.6 & 88.2 & 39.0 & 40.5
            & 22.8 & 43.5 & 21.5 & 94.7 \\

            RotTrans~\cite{chen2022rotation} & 2022
            & \underline{41.4} & \underline{99.2} & \underline{41.6} & \underline{42.4}
            & \underline{30.7} & \underline{43.2} & \underline{30.7} & \underline{97.4}  \\

            PASS~\cite{zhu2022pass} & 2022
            & 36.8 & 99.3 & 37.7 & 40.1
            & 26.7 & 45.0 & 23.1 & 93.0 \\

            SBS~\cite{he2023fastreid} & 2023
            & 14.1 & 63.3 & 18.9 & 19.8
            & 0.8 & 1.1 & 0.9 & 0.9 \\

            DCT~\cite{li2023dc} & 2023
            & 35.6 & 88.7 & 36.0 & 39.1
            & 18.5 & 40.0 & 16.0 & 90.2 \\

            VDT~\cite{Zhang_2024_CVPR} & 2024
            & 37.9 & 97.9 & 39.1 & 43.7
            & 18.1 & 43.2 & 17.8 & 92.9 \\

            DCAC~\cite{li2025unleashing} & 2025
            & 4.0 & 28.6 & 3.8 & 38.9
            & 14.3 & 33.4 & 13.0 & 75.3 \\

            \midrule

            \textbf{3D-LENS} & \textbf{2026}
            & \textbf{51.4} & \textbf{100.0} & \textbf{52.2} & \textbf{55.6}
            & \textbf{32.1} & \textbf{57.9} & \textbf{32.9} & \textbf{99.0} \\

            \bottomrule
        \end{tabular}
    \end{table}

    \begin{table}[tbp]
        \caption{
            \textbf{Ablation study on AG-ReID Ground train set.}
            We analyze the contribution of each proposed component: \textbf{Nv.} (Novel-view synthesis), \textbf{Ba.} (Background compositing), \textbf{St.} (Style transfer), \textbf{Sa.} (Balanced Sampling), \textbf{Do.} (Domain loss), and \textbf{Cu.} (Curriculum learning).
            Results are reported as mAP and Rank-1 (\%), showing the absolute value followed by the difference relative to the baseline (Row 0) in {\color{gain}green (+)} or {\color{loss}red (-)}.
            The checkmark ($\checkmark$) denotes the inclusion of the module.
            Best results are highlighted in \textbf{bold}.
        }

        \scriptsize
        \setlength{\tabcolsep}{2.5pt}

        \label{tab:ablation_ground_full}
        \centering
        \begin{tabular}{c cccccc cccc}
            \toprule
            \multicolumn{1}{c}{\textbf{Index}}
            & \multicolumn{6}{c}{\textbf{Components}}
            & \multicolumn{4}{c}{\textbf{AG-ReID Ground}} \\

            \cmidrule(lr){2-7} \cmidrule(lr){8-11}

            & \multirow{2.5}{*}{Nv.}
            & \multirow{2.5}{*}{Ba.}
            & \multirow{2.5}{*}{St.}
            & \multirow{2.5}{*}{Sa.}
            & \multirow{2.5}{*}{Do.}
            & \multirow{2.5}{*}{Cu.}
            & \multicolumn{2}{c}{$A \rightarrow G$}
            & \multicolumn{2}{c}{$G \rightarrow A$} \\

            \cmidrule(lr){8-9} \cmidrule(lr){10-11}

            & & & & & & & mAP & R-1 & mAP & R-1 \\
            \midrule

            0 & & & & & &
            & 46.9 & 54.6 & 49.8 & 60.1 \\

            \addlinespace[3pt]

            1 & \checkmark & &
            & & &
            & \resNeg{43.7}{3.2} & \resNeg{53.7}{0.9} & \resNeg{48.3}{1.5} & \resNeg{59.5}{0.6} \\

            2 & \checkmark & \checkmark &
            & & &
            & \resNeg{43.5}{3.4} & \resNeg{53.7}{0.9} & \resPos{50.2}{0.4} & \resNeg{59.3}{0.8} \\

            3 & \checkmark & & \checkmark
            & & &
            & \resPos{48.6}{1.7} & \resPos{58.6}{4.0} & \resPos{51.0}{1.2} & \resPos{62.9}{2.8} \\

            4 & \checkmark & \checkmark & \checkmark
            & & &
            & \resPos{50.9}{4.0} & \resPos{61.9}{7.3} & \resPos{51.6}{1.8} & \resPos{63.6}{3.5} \\

            5 & \checkmark & &
            & \checkmark & &
            & \resPos{47.0}{0.1} & \resPos{57.2}{2.6} & \resPos{51.7}{1.9} & \resPos{62.9}{2.8} \\

            6 & \checkmark & \checkmark & \checkmark
            & \checkmark & &
            & \resPos{51.2}{4.3} & \resPos{62.5}{7.9} & \resPos{52.8}{3.0} & \resPos{65.5}{5.4} \\

            7 & \checkmark & \checkmark & \checkmark
            & & \checkmark &
            & \resPos{51.4}{4.5} & \resPos{62.1}{7.5} & \resPos{52.8}{3.0} & \resPos{64.4}{4.3} \\

            8 & \checkmark & \checkmark & \checkmark
            & & & \checkmark
            & \resPos{51.7}{4.8} & \resPos{63.1}{8.5} & \resPos{53.1}{3.3} & \resPos{64.0}{3.9} \\

            \addlinespace[3pt]

            9 & \checkmark & \checkmark & \checkmark
            & \checkmark & \checkmark & \checkmark
            & \resPosBold{53.3}{6.4} & \resPosBold{65.2}{10.6} & \resPosBold{55.1}{5.3} & \resPosBold{67.4}{7.3} \\

            \bottomrule
        \end{tabular}
    \end{table}

    Building upon these visually aligned novel views, our robust representation learning scheme further enhances performance.
    The independent and combined additions of balanced sampling, domain loss, and curriculum learning consistently improve retrieval accuracy by preventing synthetic overfitting and stabilizing early optimization.

    Ultimately, the complete 3D-LENS framework achieves the best overall performance, delivering baseline improvements of $+6.4$ mAP pp on $A \rightarrow G$ and $+5.3$ mAP pp on $G \rightarrow A$, fully validating our approach.

    We provide qualitative results of successful novel views for AG-ReID, AG-ReID.v2, and MOO in the supplementary material.
    We also discuss and provide qualitative results for failure cases of our method in the supplementary material.

    \section{Conclusion}
    \label{sec:conclusion}

    We introduced SV AG-ReID, a setting where models must re-identify individuals across aerial and ground viewpoints without any paired cross-view training data.

    \textbf{3D-LENS} addresses this by leveraging large-scale 3D reconstruction from single-view images and rendering consistent novel viewpoints, avoiding the geometric artifacts of 2D generative models and the category restrictions of 3D templates.
    A balanced sampling strategy and curriculum learning then close the gap between synthetic and real distributions, letting the model train on views it has never directly observed.
    On AG-ReID, AG-ReID.v2, and MOO, the method consistently outperforms existing baselines, showing that class-agnostic 3D reconstruction is a viable path to cross-view re-identification without target-domain supervision.

    Looking forward, we expect that open-world 3D reconstruction will drive controllable data generation for broader generalization tasks in ReID.
    However, the reliability of such generated data remains an open challenge.
    The quality of the synthesized views is inherently bounded by the source images: low resolution, heavy occlusion, and multi-person overlap can all propagate through the reconstruction and rendering pipeline, yielding artifacts that degrade rather than enhance training.
    While synthetic data offer a scalable complement to real-world samples, they cannot yet be considered equally reliable substitutes.
    Future work should therefore focus on automated quality-control mechanisms capable of filtering or weighting synthetic views during training, ensuring that the augmented data consistently contributes to model performance.

    \subsubsection*{Acknowledgement}

    This publication was made possible by the use of the FactoryIA supercomputer, financially supported by the Ile-De-France Regional Council.

    \bibliographystyle{splncs04}
    \bibliography{main}
\includepdf[pages=-]{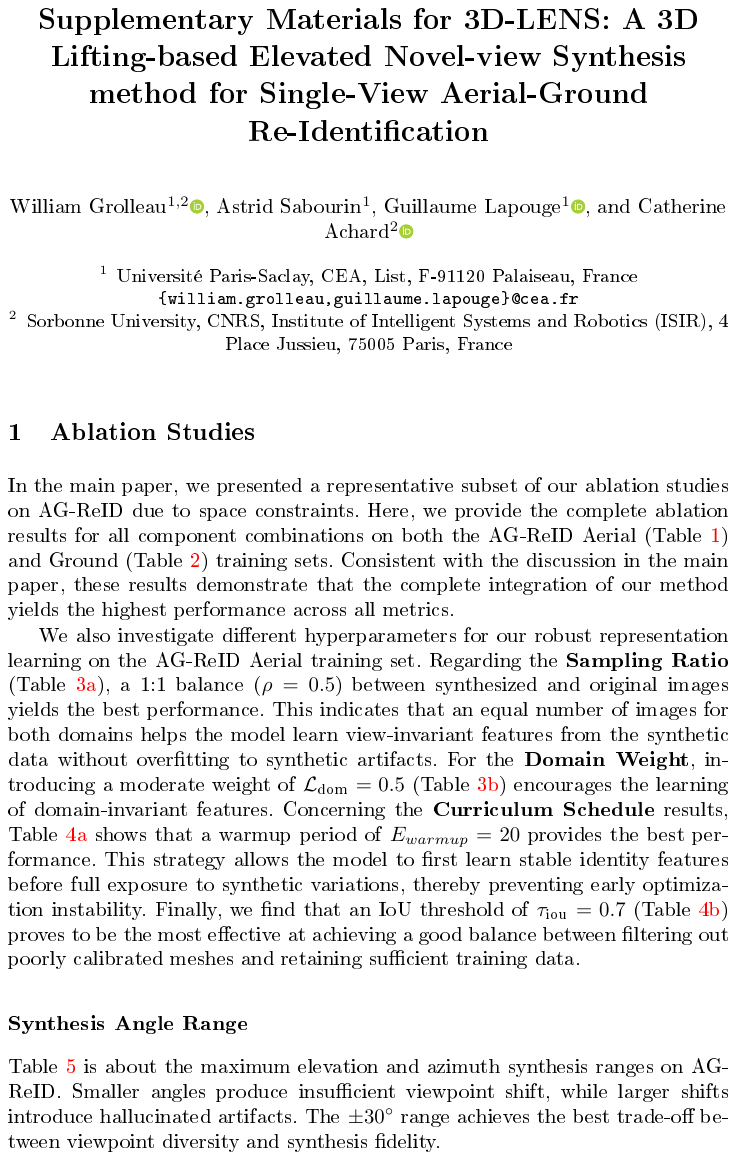}
\end{document}